\documentclass[letterpaper, 10 pt, conference]{ieeeconf}  % Comment this line out if you need a4paper
\usepackage{amsmath}
\usepackage{amssymb}
\usepackage{hyperref}
\usepackage{svg}
\usepackage[outdir = ./]{epstopdf}
\usepackage{bm}
\usepackage{derivative}
\usepackage{subcaption}
\usepackage{tikz}
\tikzset{every picture/.style={line width=0.75pt}} %set default line width to 0.75pt  
\usepackage{dblfloatfix}
\usepackage{balance}
\allowdisplaybreaks

\IEEEoverridecommandlockouts                              % This command is only needed if 
\newcommand*\dif{\mathop{}\!\mathrm{d}}
\newcommand\tth{^\text{th}}

\title{\LARGE \bf
A Rapid Trajectory Optimization and Control Framework for Resource-Constrained Applications
}

\author{Deep Parikh$^{1}$, Thomas L. Ahrens$^{1}$, Manoranjan Majji$^{2}$% <-this % stops a space
\thanks{*This work is supported by the Air Force Office of Scientific Research (AFOSR), as a part of the
SURI on OSAM project “Breaking the Launch Once Use Once Paradigm” (Grant No: FA9550-
22-1-0093).}% <-this % stops a space
\thanks{$^{1}$PhD Student, {\tt\footnotesize \{deep,tahrens\}@tamu.edu}}%
\thanks{$^{2}$Director, LASR Laboratory, Professor}%
\thanks{Department of Aerospace Engineering, Texas A\&M University, College Station, Texas, USA.}
}

\begin{document}

\maketitle
\thispagestyle{empty}
\pagestyle{empty}

%%%%%%%%%%%%%%%%%%%%%%%%%%%%%%%%%%%%%%%%%%%%%%%%%%%%%%%%%%%%%%%%%%%%%%%%%%%%%%%%
\begin{abstract}
This paper presents a computationally efficient model predictive control formulation that uses an integral Chebyshev collocation method to enable rapid operations of autonomous agents. By posing the finite-horizon optimal control problem and recursive re-evaluation of the optimal trajectories, minimization of the L2 norms of the state and control errors are transcribed into a quadratic program. Control and state variable constraints are parameterized using Chebyshev polynomials and are accommodated in the optimal trajectory generation programs to incorporate the actuator limits and keep-out constraints. Differentiable collision detection of polytopes is leveraged for optimal collision avoidance. Results obtained from the collocation methods are benchmarked against the existing approaches on an edge computer to outline the performance improvements. Finally, collaborative control scenarios involving multi-agent space systems are considered to demonstrate the technical merits of the proposed work.
\end{abstract}
%%%%%%%%%%%%%%%%%%%%%%%%%%%%%%%%%%%%%%%%%%%%%%%%%%%%%%%%%%%%%%%%%%%%%%%%%%%%%%%%
\section{Introduction}
There has been a significant leap during the past few years in satellite missions relying on in-space autonomy \cite{future_iso}. Most of these missions employ state-of-the-art Rendezvous, Proximity Operations and Docking (RPOD) and satellite swarming technologies to enable space science, earth observation, planetary defense, satellite servicing and on-orbit life extension \cite{sanchez2018starling1,adams2019double,redd2020bringing}. One of the earliest satellite servicing missions utilized a robotic manipulator to demonstrate the on-orbit capabilities in low Earth orbit \cite{ogilvie2008autonomous}, and most of the proposed missions to service and repair satellites hope to leverage manipulators to grasp the target spacecraft \cite{VISENTIN199845}.

However, the additional degrees of freedom (DoF) associated with the floating base of the robot manipulator renders the analysis of system dynamics and real-time control challenging \cite{kin_ft_base}. Consequently, most of such missions have been considered to be at a very low Technology Readiness Level (TRL) due to limited onboard autonomy constrained by Size, Weight, and Power (SWaP) \cite{knight2024advances}. On the contrary, the recent advancements in consumer electronics and sensor technology have pushed the boundaries of autonomous in-space operations \cite{doi:10.2514/6.2024-1067}. The process of qualification, verification, and validation of such novel hardware components and software packages have been greatly accelerated by the ease of availability of CubeSat manufacturers and launch providers \cite{yost2024state}. 

The Model-Predictive-Control (MPC) framework has been extensively studied for automatic control of industrial processes \cite{qin1997overview}. Although the framework has seen limited use in aerospace vehicles due to its stringent qualification, robustness and real-time requirements \cite{eren2017model}. There is an unequivocal pursuit to bridge the gap between computationally intensive MPC and resource-constrained computers by exploiting the control structures to accelerate and compress solver algorithms \cite{tinympc}. This has also motivated studies to assess the computational requirements of a pragmatic MPC-based RPOD algorithm \cite{doi:10.2514/1.G007523}.

Alongside the advances in available onboard hardware, many new computationally efficient algorithms have been developed to solve optimal control problems. In particular, pseudospectral methods are well-known for their accuracy and robustness \cite{betts1998survey}. Recent work using integral collocation, rather than the typical derivative methods, yields enhanced precision and significant runtime speedups \cite{Bai, Peck:2023}. One of these methods, known as Integral Chebyshev Collocation (ICC) \cite{Peck:2023}, has been shown to be especially accurate, and does not require the use of an another (typically Legendre) interpolating polynomial.

To that end, this paper presents a pseudospectral-based MPC framework called MPC$^3$: \textit{Model Predictive Control via Chebyshev Collocation}. The main contribution of the paper, constrained quadratic programming optimization via orthogonal approximation using ICC, is introduced in Sec.~\ref{sec:MPC3}. Performance results for the proposed method are compared with existing solvers in Sec.~\ref{sec:perf}. Finally, Sec.~\ref{sec:TPODS_DOCK} demonstrates the effectiveness of the framework to achieve the desired objective with an application of multi-state guidance logic for a docking scenario of small satellites.

\section{MPC$^3$ Formulation} \label{sec:MPC3}
\subsection{Integral Chebyshev Collocation}
Orthogonal polynomials play an important role in the fields of numerical analysis, approximation, and optimal control. A particularly attractive family of these orthogonal functions are the Chebyshev polynomials,
\begin{align} \label{eq:chebyshev}
    T_n(\tau) &= \cos \left( n \arccos \tau \right), \qquad \tau \in [-1,1]
\end{align}
which have been the subject of much work for solving initial value problems (IVPs) and boundary value problems (BVPs). Bai and Junkins developed Modified Chebyshev Picard Iteration (MCPI) to converge toward a solution for IVPs and BVPs \cite{Bai, BaiJunkins}. More recently, Peck and Majji proposed a concise method called Integral Chebyshev Collocation (ICC) \cite{Peck:2023}, the chosen framework for this paper, described briefly hereafter for a generic second-order system.

The ICC procedure begins by transforming the problem from the time domain $t \in [t_0, t_f]$ to the computational domain $\tau \in [-1,1]$ over which the Chebyshev polynomials are defined. This is achieved with the linear transformation
\begin{align} \label{eq:timetrans}
    \tau &= \frac{2t - (t_f + t_0)}{\Delta t}, \qquad \Delta t = t_f - t_0
\end{align}
The dynamics, derivative, state, and corresponding initial conditions are then cast to the computational domain as
\begin{subequations}
\begin{align}
    x^{\prime\prime}(\tau) &= \left( \frac{\Delta t}{2} \right)^2 \ddot{x}(t) \\
    x^{\prime}(\tau) &= \left( \frac{\Delta t}{2} \right) \dot{x}(t), \quad x^{\prime}(-1) = \left( \frac{\Delta t}{2} \right) \dot{x}_0 \\
    x(\tau) &= x(t), \quad x(-1) = x_0
\end{align}
\end{subequations}

Now that the problem is represented in the Chebyshev domain, an $n\tth$ order series of Chebyshev polynomials is used to represent the dynamics,
\begin{align} \label{eq:xpptau}
    x^{\prime\prime}(\tau) &= \sum_{i=0}^{n} T_i(\tau) \alpha_i = \bm{T}(\tau) \bm{\alpha}
\end{align}
where $\bm{T}(\tau) = [ T_0(\tau), T_1(\tau), \ldots, T_n(\tau) ] \in \mathbb{R}^{1 \times (n+1)}$ is a row vector of Chebyshev polynomials evaluated at $\tau$ and $\bm{\alpha} = [ \alpha_0, \alpha_1, \ldots, \alpha_n ]^\top \in \mathbb{R}^{(n+1) \times 1}$ is a column vector of unknown coefficients. Integrating this expression of Chebyshev polynomials once or twice yields the derivative or state expression,
\begin{align}
    x^{\prime}(\tau) &= \bm{\beta}(\tau) \bm{\alpha} + x^{\prime}(-1) \label{eq:xptau} \\
    x(\tau) &= \bm{\gamma}(\tau) \bm{\alpha} + x^{\prime}(-1) (\tau + 1) + x(-1) \label{eq:xtau}
\end{align}
where $\bm{\beta}(\tau), \bm{\gamma}(\tau) \in \mathbb{R}^{1 \times (n+1)}$ are introduced as a shorthand to denote the first and second integration operators given in \cite{Peck:2023}, respectively.

This procedure is then extended to the set of nodes $\bm{\tau} = [ \tau_0, \tau_1, \ldots, \tau_n]^\top$ that minimize the interpolation error: the Chebyshev-Gauss (CG) nodes which the roots of the $(n+1)\tth$ order Chebyshev polynomial \cite{CPNumAna_minimax},
\begin{align} \label{eq:cheb_zeros}
    \tau_{k-1} &= \cos \left( \frac{(k - \frac{1}{2}) \pi}{n+1} \right), \quad k = [1, 2, \ldots, n+1]
\end{align}
Using the CG nodes $\bm{\tau}$ leads to a vector-matrix collocation formulation,
\begin{subequations}
\begin{align}
    \bm{x}^{\prime\prime}(\bm{\tau}) &= \mathcal{T}(\bm{\tau}) \bm {\alpha} \label{eq:xppCG} \\
    \bm{x}^{\prime}(\bm{\tau}) &= \bm{\beta}(\bm{\tau}) \bm{\alpha} + \bm{x}^{\prime}(-1) \label{eq:xpCG} \\
    \bm{x}(\bm{\tau}) &= \bm{\gamma}(\bm{\tau}) \bm{\alpha} + \bm{x}^{\prime}(-1) (\bm{\tau} + 1) + \bm{x}(-1) \label{eq:xCG}
\end{align}
\end{subequations}
where $\mathcal{T}(\bm{\tau}), \bm{\beta}(\bm{\tau}), \bm{\gamma}(\bm{\tau}) \in \mathbb{R}^{(n+1) \times (n+1)}$ are matrix versions of the previously defined vectors. Once the approximation order $n$ is chosen, these matrices are constant and can be computed {\it a priori}, leading to a drastic improvement in computational efficiency. To solve a linear system, the coefficients can be found via a single matrix inversion, while nonlinear systems require an iterative method.

\begin{table*}[b!]
    \centering
    \caption{Comparison of Discrete MPC and MPC$^3$}
    \label{tab:compare_formulation}
    {\renewcommand{\arraystretch}{1.5}    
    \begin{tabular}{|c|c|c|}
    \cline{2-3}
    \multicolumn{1}{c|}{} & MPC & MPC$^3$ \\ 
    \hline State Equation  & $\bm{x}_{k+1} = \boldsymbol{A}_d\boldsymbol{x}_k + \boldsymbol{B}_d\boldsymbol{u}$ , $\bm{u} \in \mathbb{R}^{m}$, $\bm{x} \in \mathbb{R}^{q}$ & $\dot{\bm{x}}(t) = \bm{g}(\bm{t},\bm{x},\bm{u})$, $\bm{u} \in \mathbb{R}^{m}$, $\bm{x} \in \mathbb{R}^{q}$\\
         \hline
       Recursion  & $\boldsymbol{y}  = \boldsymbol{S}_x \boldsymbol{x}_k + \boldsymbol{S}_y \boldsymbol{u}$  & $\begin{matrix} 
    x^{\prime\prime}(\tau) = u(\tau) = \bm{T}(\tau) \bm{\alpha}  \\
    x^{\prime}(\tau) = \bm{\beta}(\tau) \bm{\alpha} + x^{\prime}(-1) \\
    x(\tau) = \bm{\gamma}(\tau) \bm{\alpha} + x^{\prime}(-1)(\tau + 1) + x(-1) 
    \end{matrix}$  \\
         \hline
       Objective function  & $\displaystyle\sum_{i=0}^{p-1} \Big(\boldsymbol{u}^T_{k+i}\boldsymbol{W}_u\boldsymbol{u}_{k+i}\ +  \left(\boldsymbol{y}_{k+i}-\boldsymbol{y}_r\right)^T\boldsymbol{W}_y\left(\boldsymbol{y}_{k+i}-\boldsymbol{y}_r\right)\Big)$ & $\displaystyle\int_{t_0}^{t_f} \left[ \bm{u}^\top \bm{W_u} \bm{u} + (\bm{x} - \bm{x}_t)^\top \bm{W_x} (\bm{x} - \bm{x}_t) \right] \dif{t}$\\
         \hline
       Decision Variables  & $\bm{u_k} \in \mathbb{R}^{p \times m}$ & $\chi = \left[ \bm{\alpha}, \varepsilon \right]^\top \in \mathbb{R}^{(q(n+1)+1) \times 1}$ \\
         \hline
       \# of Equality Constraints  & $0$ & $2(n+2) \times q$\\
         \hline
       \# of Inequality Constraints  & $2p \times q$ & $4(n+1) \times q$\\
         \hline
    \end{tabular}}
\end{table*}

\subsection{$L_2$ Minimization}
The objective of a typical MPC is to solve an optimization problem at each time instance for a specified control horizon and produce a control input that minimizes the objective function \cite{rawlings2017model}. Unlike traditional discrete MPC, the cost functional for the linear dynamics $\dot{\bm{x}}(t) = \bm{g}(\bm{t},\bm{x},\bm{u})$ is given as a continuous integral from the current time $t_0$ to some future time $t_f$,
\begin{align}
    J &= \int_{t_0}^{t_f} \left[ \bm{u}^\top \bm{W_u} \bm{u} + (\bm{x} - \bm{x}_t)^\top \bm{W_x} (\bm{x} - \bm{x}_t) \right] \dif{t}
\end{align}
where $\bm{u} \in \mathbb{R}^{m}$ is the control, $\bm{x}, \bm{x}_t \in \mathbb{R}^{q}$ are the instantaneous and target state, and $\bm{W_u} \in \mathbb{R}^{m \times m}$ and $\bm{W_x} \in \mathbb{R}^{q \times q}$ are the diagonal weight matrices for the control and states, respectively. The time or prediction horizon associated with the problem is now $\Delta t$ from Eq.~\eqref{eq:timetrans}.

Hereafter, we consider problems which can be written or linearized in the form of a double integrator, $\bm{\ddot{x}}(t) = \bm{u}(t)$, the benefit of which is that both the states and controls can be represented with the same set of unknown polynomial coefficients. For the sake of brevity, a 1-DoF double integrator will be used in this derivation, but the procedure easily extends to multi-DoF problems by using another set of coefficients for each additional state, as is done in Sec.~\ref{sec:TPODS_DOCK}. The cost functional is then converted to the Chebyshev domain as
\begin{align}
    J &= \frac{\Delta t}{2} \int_{-1}^{1} \big[ u(\tau) W_u u(\tau) + (x(\tau) - x_t) W_x (x(\tau) - x_t) \nonumber \\
    &\qquad \quad \qquad + (x^{\prime}(\tau) - x_t^{\prime}) W_{x^\prime}  (x^{\prime}(\tau) - x_t^{\prime}) \big] \dif{\tau}
\end{align}
where the integrand includes terms with quadratic, linear, and no dependence on the coefficients $\bm{\alpha}$. The terms independent of the decision variable are constant, and can be removed from the minimization problem \cite{borrelli2017predictive}, but the remaining integrand expression is not well-posed given the Chebyshev polynomial representations of the state and derivative.

To evaluate this integral, a Gaussian quadrature rule is applied to approximate the integral,
%\cite{GaussQuad},
\begin{align} \label{eq:GaussQuad}
    \int_{-1}^{-1} g(\tau) \dif{\tau} &\simeq \sum_{i=0}^{n} w_i g(\tau_i)
\end{align}
where $w_i$ are the quadrature weights \cite{Notaris} and $g(\tau_i)$ are the integrand evaluated at the CG nodes. Thus, the problem is written in the typical quadratic problem (QP) form
\begin{align} \label{eq:qp_form}
    \min J &= \frac{1}{2} \bm{\alpha}^\top H \bm{\alpha} + \bm{f}^\top \bm{\alpha}
\end{align}
where the matrix $H \succ 0$ and vector $\bm{f}$ are given as
\begin{subequations}
\begin{align}
    H &= \Delta t \Bigg[ \sum_{i=0}^{n} w_i \left( (\bm{T}(\tau_i))^\top W_u \bm{T}(\tau_i) + (\bm{\gamma}(\tau_i))^\top W_x \bm{\gamma}(\tau_i) \right. \nonumber \\
    &\qquad \quad \left. + (\bm{\beta}(\tau_i))^\top W_{x^\prime} \bm{\beta}(\tau_i) \right) \Bigg] \in \mathbb{R}^{(n+1) \times (n+1)} \label{eq:H} \\
    \bm{f} &= \Delta t \Bigg[ \sum_{i=0}^{n} w_i \left( W_x \bm{\gamma}(\tau_i) \left(  x^\prime(-1)(\tau_i + 1) + x(-1) - x_t \right) \right. \nonumber \\
    &\qquad \quad \left. + W_{x^\prime} \bm{\beta}(\tau_i) \left( x^\prime(-1) - x_t^\prime \right) \right) \Bigg] \in \mathbb{R}^{(n+1) \times 1} \label{eq:f}
\end{align}
\end{subequations}

\subsection{State and Input Constraints}
For most practical applications, the control inputs are limited to a predefined threshold, which can be incorporated into the optimal control framework by the addition of a slack variable and input equality constraints. Similarly, it is often desired to restrict certain states throughout the duration of the trajectory within a specific boundary. For example, limiting approach velocity to a target can still allow for an overall aggressive position controller without any undesirable overshoot. Again, this can be achieved by the method of a slack variable and inequality constraints. Finally, equality constraints are placed on the states at the initial time to ensure the initial boundary conditions are met.

With these modifications, the QP of Eq.~\eqref{eq:qp_form} is written with equality and inequality constraints as
\begin{align} \label{eq:qp_slack}
\begin{split}
    \min_{\bm{\chi}} J &= \frac{1}{2} \bm{\chi}^\top H \bm{\chi} + \bm{f}^\top \bm{\chi} \\
    \text{s.t.} \quad A_\text{eq} \bm{\chi} &= \mathbf{b}_\text{eq}, \quad A \bm{\chi} \leq \mathbf{b}
\end{split}
\end{align}
where, with some abuse of notation,
\begin{subequations}
\begin{align}
    \bm{\chi} &= \begin{bmatrix} \bm{\alpha} \\ \varepsilon \end{bmatrix} \in \mathbb{R}^{(n+2) \times 1} \\
    H &= \begin{bmatrix} H & \bm{0}_{n+1,1} \\ \bm{0}_{1,n+1} & \rho \end{bmatrix} \in \mathbb{R}^{(n+2) \times (n+2)} \\
    \bm{f} &= \begin{bmatrix} \bm{f} \\ 0 \end{bmatrix} \in \mathbb{R}^{(n+2) \times 1} \\
    A_\text{eq} &= \begin{bmatrix} \bm{\gamma}(-1) & 0 \\ \bm{\beta}(-1) & 0 \end{bmatrix} \in \mathbb{R}^{2 \times (n+2)} \\
    \mathbf{b}_\text{eq} &= \begin{bmatrix} 0 \\ 0 \end{bmatrix} \in \mathbb{R}^{2 \times 1} \\
    A &= \begin{bmatrix} \mathcal{T}(\bm{\tau}) & -V_u \\ -\mathcal{T}(\bm{\tau}) & -V_u  \\ \bm{\beta}(\bm{\tau}) & -V_{x^\prime} \\ -\bm{\beta}(\bm{\tau}) & -V_{x^\prime} \end{bmatrix} \in \mathbb{R}^{4(n+1) \times (n+2)} \\
    \mathbf{b} &= \begin{bmatrix} u_\text{max}^\prime \\ -u_\text{min}^\prime \\ x_\text{max}^\prime - x^\prime(-1) \\ -x_\text{min}^\prime + x^\prime(-1) \end{bmatrix} \in \mathbb{R}^{4(n+1) \times 1}
\end{align}
\end{subequations}
Here, $\rho$ is the weight factor associated with the slack variable $\varepsilon$, $u_\text{max}^\prime/u_\text{min}^\prime$ and $x_\text{max}^\prime/x_\text{min}^\prime$ are the constraints on the control and velocity, and $V_u$ and $V_{x^\prime}$ are used to regulate the softness of the constraints. A comprehensive comparison of MPC$^3$ with discrete MPC along with dimensions of the optimization program is given in Table~\ref{tab:compare_formulation}.

\section{Performance Comparison} \label{sec:perf}
The constrained optimization problem, now formulated in the computational domain as a QP, is solved with MATLAB\textsuperscript{\textregistered}'s \textit{quadprog} using the active-set algorithm and warm-start enabled. Importantly, note that of the matrices and vectors defined in the previous section, $H$, $A_\text{eq}$, $\mathbf{b}_\text{eq}$, and $A$ remain unchanged, regardless of time horizon or initial conditions. Only $\bm{f}$ and $\mathbf{b}$ need to be updated at each step, resulting in an efficient formulation. The coefficients and slack variable are found by minimizing the objective function of Eq.~\eqref{eq:qp_slack}, and the instantaneous control is calculated as $u(-1) = \bm{T}(-1) \bm{\alpha}$. ICC, or any other integrator, can then be used to propagate the true (generally nonlinear) dynamics until reaching the next control input time. 

\begin{figure}[t]
    \center
    \includegraphics[width=0.48\textwidth]{figures/double_int_compare.eps}
    \caption{Comparison of trajectories and optimal input for double integrator for $T_s = 0.5$~s, $p = 5$ samples, and $n = 3$.}
    \label{fig:compare_1DOF}
\end{figure}

\begin{figure*}[b]  
     \begin{subfigure}[b]{0.245\textwidth}
        \centering
         \includegraphics[width=\textwidth]{figures/cpu_time_DOF.eps}
         \caption{Run time for different $q$ ($p = 5$)}
         \label{fig:perf_a}
     \end{subfigure}
     \centering  
     \begin{subfigure}[b]{0.245\textwidth}
        \centering
         \includegraphics[width=\textwidth]{figures/memory_DOF.eps}
         \caption{Memory for different $q$ ($p = 5$)}
         \label{fig:perf_b}
    \end{subfigure}
     \centering    
    \begin{subfigure}[b]{0.235\textwidth}
        \centering
         \includegraphics[width=\textwidth]{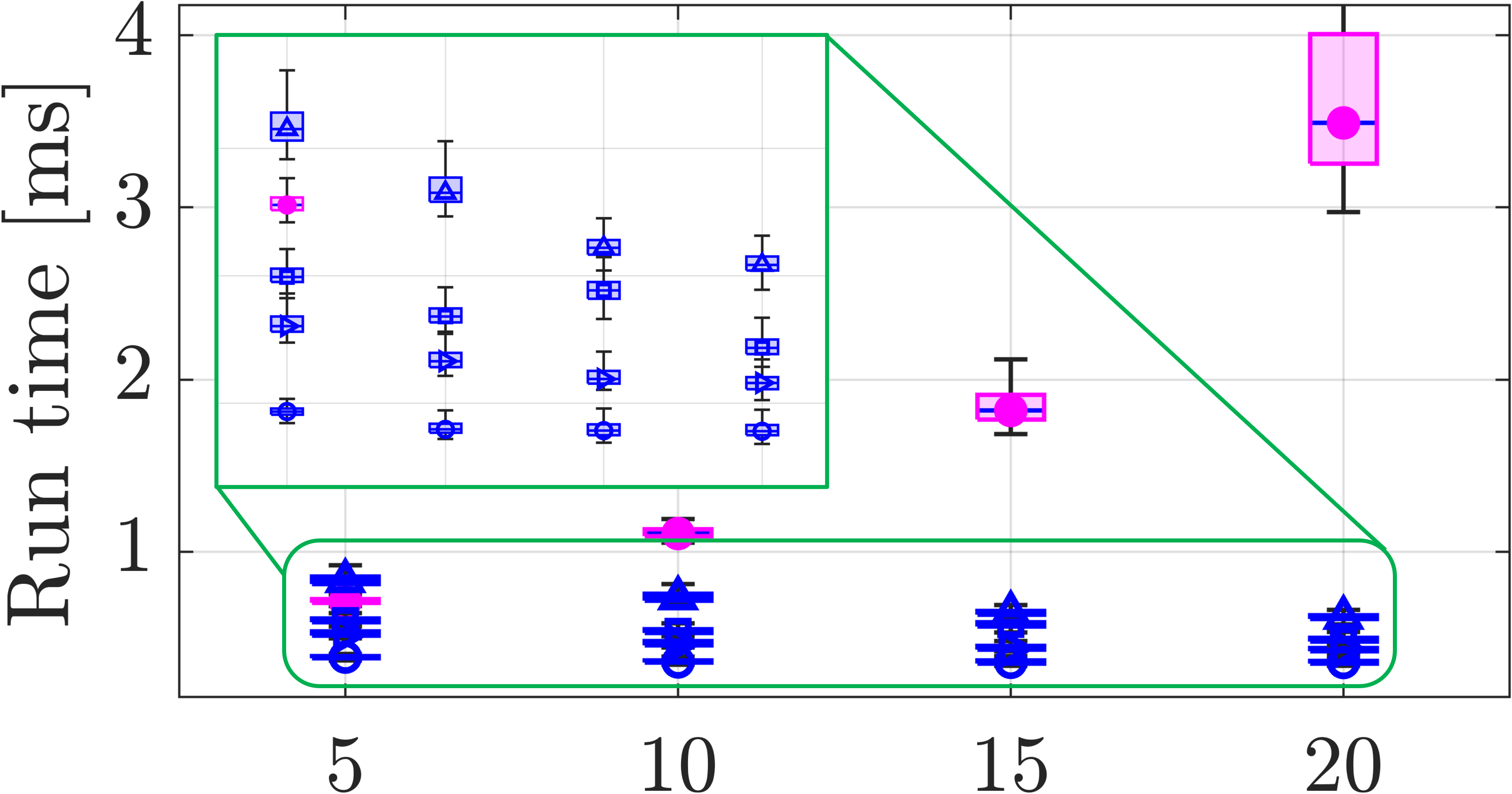}
         \caption{Run time for different $p$ ($q = 6$)}
         \label{fig:perf_c}
    \end{subfigure}
    \centering 
    \begin{subfigure}[b]{0.25\textwidth}
        \centering
         \includegraphics[width=\textwidth]{figures/memory_p.eps}
         \caption{Memory for different $p$ ($q = 6$)}
         \label{fig:perf_d}
    \end{subfigure}
    \caption{Computational performance of MPC$^3$ for different state dimensions and control horizon. \textbf{Generated on a computer} with $11\tth$ Gen Intel\textsuperscript{\textregistered} Core\textsuperscript{\texttrademark} \ i5-1135G7 @ 2.40GHz, 16 GB RAM, for running a MATLAB\textsuperscript{\textregistered} script. MATLAB\textsuperscript{\textregistered} interface for OSQP is used to integrate OSQP \cite{osqp-codegen}.}
    \label{fig:perf_comp}
\end{figure*}

\begin{figure*}[b]  
     \begin{subfigure}[b]{0.245\textwidth}
        \centering
         \includegraphics[width=\textwidth]{figures/cpu_time_DOF_teen.eps}
         \caption{Run time for different $q$ ($p = 5$)}
     \end{subfigure}
     \centering  
     \begin{subfigure}[b]{0.245\textwidth}
        \centering
         \includegraphics[width=\textwidth]{figures/memory_DOF_teen.eps}
         \caption{Memory for different $q$ ($p = 5$)}
    \end{subfigure}
     \centering    
    \begin{subfigure}[b]{0.245\textwidth}
        \centering
         \includegraphics[width=\textwidth]{figures/cpu_time_p_teen.eps}
         \caption{Run time for different $p$ ($q = 6$)}
    \end{subfigure}
    \centering 
    \begin{subfigure}[b]{0.245\textwidth}
        \centering
         \includegraphics[width=\textwidth]{figures/memory_p_teen.eps}
         \caption{Memory for different $p$ ($q = 6$)}
    \end{subfigure}
    \caption{Computational performance of MPC$^3$ for different state dimensions and control horizon, with fixed n = 3. \textbf{Generated on Teensy 4.1 development board}, for running a single iteration of optimization routine, built and deployed from a MATLAB\textsuperscript{\textregistered} script.}
    \label{fig:perf_comp_teen}
\end{figure*}

A comparison of the optimal trajectory solutions for a double integrator system obtained via a conventional MPC and MPC$^3$ is presented in Fig.~\ref{fig:compare_1DOF}. Although both trajectories are very close to one another, it is important to note that both formulations do not have the same weighing factors $\boldsymbol{W_u}$ and $\boldsymbol{W_x}$, and turning of these parameters is often required to ensure both solutions are identical. 

\subsection{Variation with State Dimension}
To establish the computational gains of the proposed formulation, extensive simulation studies have been conducted and their results are presented in Fig.~\ref{fig:perf_comp}. For the first analysis, the control horizon has been kept fixed at $p=5$ samples and the dimension of the state $q$ has been varied. The average run time and memory requirement  of a single optimization routine is recorded and the comparison of MPC$^3$ with various solvers is presented in Figs.~\ref{fig:perf_a} and~\ref{fig:perf_b}. Another benefit of this MPC$^3$ formulation is that previous solutions can be used to hot start the current iteration. In particular, if the time between control inputs is relatively small and the trajectory remains relatively similar on this scale (as is expected with MPC), the previous free variables $\bm{\chi}$ can be used as an initial guess for the current coefficients and slack variable, significantly reducing the time to solve QP. 

\subsection{Variation with Control Horizon}
Further, the state dimension $q=6$ has been kept unchanged and the control horizon has been varied from $p=5$ to $p=20$ with an increment of $5$. The results presented in Figs.~\ref{fig:perf_c} and~\ref{fig:perf_d} particularly underscore the benefits of MPC$^3$. Since the ICC solution is global over the time horizon of interest, the runtime or memory requirements of MPC$^3$ formulation do not increase with a larger time horizon. While a larger approximation order $n$ may be necessary to capture the greater dynamic range over this time horizon, a small increase in $n$ is sufficient for these changes, which is shown to only marginally increase runtime.

\subsection{Performance Comparison on an Edge Computer}
A similar analysis is also carried out for Teensy 4.1\footnote{https://www.pjrc.com/store/teensy41.html} development board to qualify the computational benefits of the proposed framework on an embedded hardware. The MATLAB® scripts used for the earlier analysis have been built and deployed on the Teensy board using automatic code generation. The results of this exercise are presented in Figure~\ref{fig:perf_comp_teen}, follows a similar trend of run time and memory usage for various state dimensions and control horizon. Most importantly, as the control horizon increases, the RAM usage crosses the available on-board RAM of $512$~kb and the conventional MPC could no longer be implemented on the hardware. However, as mentioned earlier, the resource utilization does not increase with the control horizon within the MPC$^3$ framework. This enables autonomous applications which were deemed infeasible due to high computational cost and memory usage. 

\section{Application To Space Proximity Operations}\label{sec:TPODS_DOCK}
\begin{figure}[t!]
\centerline{\includegraphics[width=0.35\textwidth]{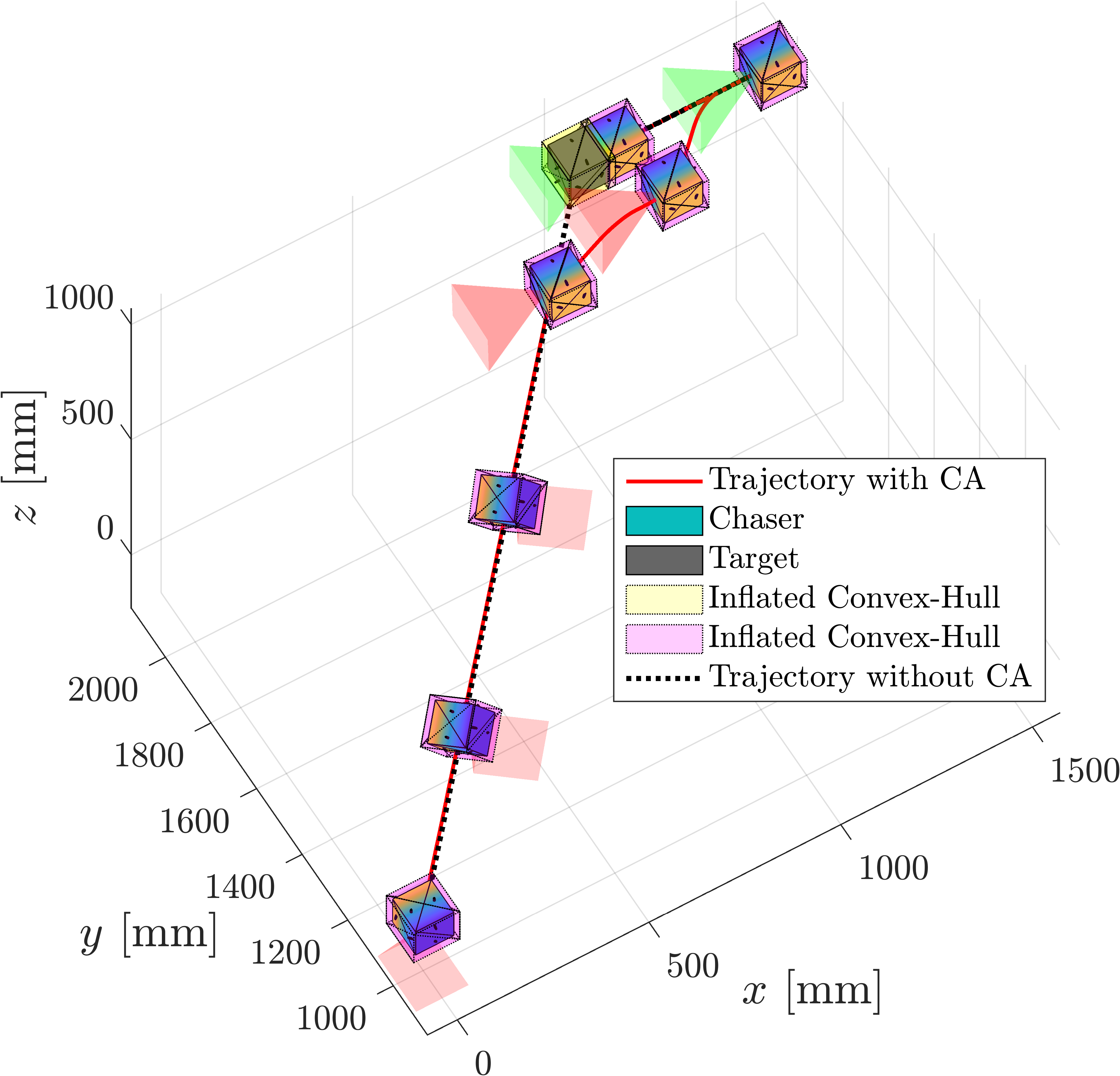}}
\caption{Guidance algorithm for docking of TPODS with a stationary target consists of three distinct guidance modes. For each mode, TPODS is commanded to move along the optimal trajectory, which passes through the target if no collision avoidance maneuver is executed. The availability of vision measurements is depicted by a color change of the FOV cone.}
\label{fig:guidance}
\end{figure}

The Transforming Proximity Operations and
Docking System (TPODS) is a conceptual 1U CubeSat module, developed by the Land, Air and Space Robotics (LASR) laboratory at
Texas A\&M University \cite{TPODS_system,TPODS_GNC24}. The proposed MPC$^3$ framework is utilized to enable robust and safe proximity operations of the TPODS module. The objective is to dock a chaser TPODS to a stationary target using a multi-state guidance algorithm. The TPODS module is equipped with Ultra-Wide-Band (UWB) radar range and monocular-vision camera sensors \cite{TPODS_GNC24}. A unified pose estimator can be leveraged to fuse the range from stationary anchors with monocular vision data to produce consistent pose estimates. The guidance logic for the docking scenarios considered in this paper is shown in Fig.~\ref{fig:guidance}.

\subsection{System Dynamics}
The UWB sensor is not mounted on the respective centers of mass for the TPODS module, resulting in a coupled rotational and translational motion of the UWB sensor relative to the stationary anchors\cite{TPODS_multi_agent}.
For the simulation studies presented in this section, the TPODS dynamics are linearized into a double integrator in three translation axes. Once the set of optimal control inputs is obtained for the selected control horizon, the control inputs at the first instance are determined and applied to the non-linear dynamics, and the true states are propagated. The current states are fed to the multi-state guidance algorithm after an appropriate process noise has been injected into the velocity states. Finally, the desired values of the attitude and position computed by the guidance algorithm are fed to the optimization routine and the process continues.

\subsection{Keep-Out-Constraint}
One of the prominent safety requirements for the majority of real-world operations is Keep-Out-Constraints (KOC). Due to inherent uncertainties associated with localizing objects, it is necessary to account for the margin of safety. A differential collision detection for convex polytopes (DCOL) can provide a robust and efficient avenue to enforce KOC \cite{DCOL}. Collision detection via DCOL works by inflating polytopic convex hulls of the target and chaser by a scaling factor $s$\cite{TPODS_multi_agent}. When $s$ is found to be greater than one, both bodies are separated. For a given set of polytopes, the minimum value of $s$ can be computed by solving a linear program with inequality constraints stemming from the fact that the intersection point is a member of both inflated polytopes.

Once the collision is detected, corrective action needs to be taken such that the chaser moves away from the target. For the analysis presented in this paper, the target is considered stationary and the attitude of the chaser is not varied during collision avoidance maneuvers. Hence, the general direction in which the separation is achieved can be inferred from the relative change of the $s$ with respect to the position of the chaser. The optimization problem given by Eq.~\eqref{eq:qp_slack} is first solved without any additional collision avoidance constraints. Next, the value of $s$ for all instances within the control horizon is computed. 

If the value of $s$ for any instance within the control horizon is less than a predefined soft limit $s_\text{thr}$, the partial derivative $\pdv{s}{\bm{r}_c}$ is computed and the optimization problem is modified\cite{TPODS_multi_agent}. The modified input sequence which ensures a collision-free trajectory is then applied to the chaser and the process is repeated until the chaser is within a specified radius of the target. As shown in Fig.~\ref{fig:guidance}, collision avoidance with DCOL results in a smooth trajectory. In addition to the least energy consumption, the collision avoidance approach based on the polytopic hulls of respective bodies prevents the optimization problem from becoming infeasible due to geometrical complexities \cite{DCOL}.  

\subsection{Optimal Trajectories for Docking}
To validate the proposed framework, the orientation of the docking face on the target and the initial pose of the chaser is selected such that the extremal trajectory of the chaser passes through the target. Individual translation velocities are restricted to $0.02$~m/s with a softness factor of $0.1$ and the control forces are restricted to $10$~mN in each axis with a softness factor of $0.01$. The soft target for scaling factor $s_\text{thr}$ is picked as $1.5$ to ensure robustness against pose estimation errors. The collision detection and avoidance is switched on when the chaser is within $0.4$~m radius of the target and switched off during the final phase of docking. The attitude controller acts independently and leverages errors in desired and current quaternion to drive the chaser to the desired orientation during various guidance modes. 

\begin{figure}[t!]
    \centering
     \includegraphics[width=0.45\textwidth]{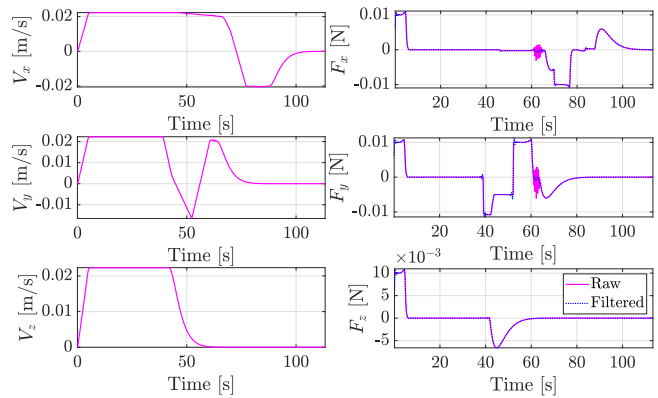}
    \caption{Time history of translation velocities and desired forces for docking scenario shown in Fig.~\ref{fig:guidance}}
    \label{fig:states}
\end{figure}

From the time histories of translation velocities presented in Fig.~\ref{fig:states}, the effectiveness of the soft constraints on the velocity can be observed. The algorithm successfully manages to restrict the velocities around the enforced soft limits. As mentioned earlier, these constraints can be enforced more aggressively using respective softness parameters and their violations can be heavily penalized using the weight factor of the slack variable. 

Barring the sporadic chattering, the desired forces also remain within the specified constraints throughout the motion. The isolated oscillations increase but average out towards the true discontinuous control with approximation order and are more generally known as Gibbs phenomenon \cite{Peck:2023}. These are undesirable and can result in wear and tear of the actuators. However, a careful selection of the weight factors and intermediate band-reject filters can help eliminate the oscillations to produce smooth control commands.  

\begin{figure}[t!]
    \centering
     \includegraphics[width=0.47\textwidth]{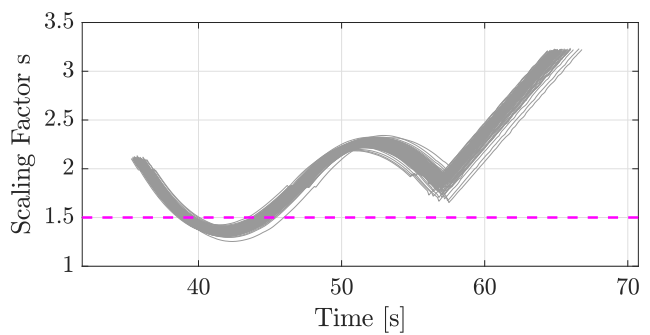}
    \caption{Scaling factor for 500 run Monte Carlo simulations.}
    \label{fig:MC_s}
\end{figure}

Finally, to assess the robustness of the collision avoidance algorithm, 500 run Monte Carlo simulations are performed. The true states of the chaser are infused with an appropriate process noise while the target position and initial position of the chaser are left unchanged. The trajectory of the scaling factor for all Monte Carlo iterations is shown in Fig.~\ref{fig:MC_s}, where the effectiveness of the collision avoidance algorithm in keeping the scaling factor near the soft threshold is evident. The softness of translation velocity and collision avoidance constraints helps in keeping the optimization problem feasible, ensuring tractable input forces.

\section{Conclusions}
A computationally efficient trajectory optimization and control framework using integral Chebyshev collocation is introduced in this paper and its performance is compared with existing optimization solvers, on a conventional and edge computers. The utility of the framework is demonstrated by formulating a model predictive controller to enable proximity operations of CubeSat agents. The state and control inequality constraints are added, along with functionality to define softness/hardness to enhance the practicality of the framework. A robust and efficient collision avoidance based on polytopic hulls makes this framework particularly attractive for the majority of the real-world applications. The current focus of the authors is to demonstrate the planar proximity operations of TPODS MK-E motion emulators using the MPC$^3$ framework and packaging for eventual release, enabling widespread adoption for resource constrained autonomous applications.

\section*{Acknowledgment}
Program monitors for the AFOSR SURI on OSAM, Dr. Andrew Sinclair  and Mr. Matthew Cleal of AFRL are gratefully acknowledged for their watchful guidance. Prof. Howie Choset of CMU, Mr. Andy Kwas of Northrop Grumman Space Systems and Prof. Rafael Fierro of UNM are acknowledged for their motivation, technical support, and discussions. Dr. Geordan Gutow of CMU is acknowledged for introducing DCOL framework the authors.


%%%%%%%%%%%%%%%%%%%%%%%%%%%%%%%%%%%%%%%%%%%%%%%%%%%%%%%%%%%%%%%%%%%%%%%%%%%%%%%%

%%%%%%%%%%%%%%%%%%%%%%%%%%%%%%%%%%%%%%%%%%%%%%%%%%%%%%%%%%%%%%%%%%%%%%%%%%%%%%%%
\balance
%%%%%%%%%%%%%%%%%%%%%%%%%%%%%%%%%%%%%%%%%%%%%%%%%%%%%%%%%%%%%%%%%%%%%%%%%%%%%%%%
\bibliographystyle{ieeetr}
\bibliography{ref}

\begin{thebibliography}{10}

\bibitem{future_iso}
E.~Rodgers, ``The future of space operations,'' in {\em AIAA SciTech Forum and Exposition}, January 2024.

\bibitem{sanchez2018starling1}
H.~Sanchez, D.~McIntosh, H.~Cannon, C.~Pires, J.~Sullivan, S.~D’Amico, and B.~O’Connor, ``Starling1: Swarm technology demonstration,'' 2018.

\bibitem{adams2019double}
E.~Adams, D.~O'Shaughnessy, M.~Reinhart, J.~John, E.~Congdon, D.~Gallagher, E.~Abel, J.~Atchison, Z.~Fletcher, M.~Chen, {\em et~al.}, ``Double asteroid redirection test: The earth strikes back,'' in {\em 2019 IEEE Aerospace Conference}, pp.~1--11, IEEE, 2019.

\bibitem{redd2020bringing}
N.~T. Redd, ``Bringing satellites back from the dead: Mission extension vehicles give defunct spacecraft a new lease on life-[news],'' {\em IEEE Spectrum}, vol.~57, no.~8, pp.~6--7, 2020.

\bibitem{ogilvie2008autonomous}
A.~Ogilvie, J.~Allport, M.~Hannah, and J.~Lymer, ``Autonomous satellite servicing using the orbital express demonstration manipulator system,'' in {\em Proc. of the 9th International Symposium on Artificial Intelligence, Robotics and Automation in Space (i-SAIRAS'08)}, pp.~25--29, 2008.

\bibitem{VISENTIN199845}
G.~Visentin and D.~Brown, ``Robotics for geostationary satellite servicing,'' {\em Robotics and Autonomous Systems}, vol.~23, no.~1, pp.~45--51, 1998.
\newblock Space Robotics in Europe.

\bibitem{kin_ft_base}
Z.~Vafa and S.~Dubowsky, ``The kinematics and dynamics of space manipulators: The virtual manipulator approach,'' {\em The International Journal of Robotics Research}, vol.~9, no.~4, pp.~3--21, 1990.

\bibitem{knight2024advances}
B.~F. Knight, ``Advances in space trusted autonomy,'' in {\em Autonomous Systems: Sensors, Processing, and Security for Ground, Air, Sea, and Space Vehicles and Infrastructure 2024}, vol.~13052, p.~1305208, SPIE, 2024.

\bibitem{doi:10.2514/6.2024-1067}
J.~Rudico, J.~T. Nichols, and G.~Rogers, {\em Potential United States Space Force Mission Life Extension Applications}.

\bibitem{yost2024state}
B.~Yost and S.~Weston, ``State-of-the-art small spacecraft technology,'' tech. rep., National Aeronautics and Space Administration, 2024.

\bibitem{qin1997overview}
S.~J. Qin and T.~A. Badgwell, ``An overview of industrial model predictive control technology,'' in {\em AIche symposium series}, vol.~93, pp.~232--256, New York, NY: American Institute of Chemical Engineers, 1971-c2002., 1997.

\bibitem{eren2017model}
U.~Eren, A.~Prach, B.~B. Ko{\c{c}}er, S.~V. Rakovi{\'c}, E.~Kayacan, and B.~A{\c{c}}{\i}kme{\c{s}}e, ``Model predictive control in aerospace systems: Current state and opportunities,'' {\em Journal of Guidance, Control, and Dynamics}, vol.~40, no.~7, pp.~1541--1566, 2017.

\bibitem{tinympc}
K.~Nguyen, S.~Schoedel, A.~Alavilli, B.~Plancher, and Z.~Manchester, ``Tinympc: Model-predictive control on resource-constrained microcontrollers,'' in {\em IEEE International Conference on Robotics and Automation (ICRA)}, 2024.

\bibitem{doi:10.2514/1.G007523}
A.~Fear and E.~G. Lightsey, ``Autonomous rendezvous and docking implementation for small satellites using model predictive control,'' {\em Journal of Guidance, Control, and Dynamics}, vol.~47, no.~3, pp.~539--547, 2024.

\bibitem{betts1998survey}
J.~T. Betts, ``Survey of numerical methods for trajectory optimization,'' {\em Journal of guidance, control, and dynamics}, vol.~21, no.~2, pp.~193--207, 1998.

\bibitem{Bai}
X.~Bai, {\em Modified Chebyshev-Picard Iteration Methods for Solution of Initial Value and Boundary Value Problems}.
\newblock PhD thesis, Texas A\&M University, 2010.

\bibitem{Peck:2023}
C.~Peck, {\em Adaptive Collocation Methods Using Chebyshev Integration}.
\newblock PhD thesis, Texas A\&M University, 2023.

\bibitem{BaiJunkins}
X.~Bai and J.~L. Junkins, ``Modified chebyshev-picard iteration methods for solution of boundary value problems,'' {\em Journal of Astronautical Sciences}, vol.~58, pp.~615--642, 2011.

\bibitem{CPNumAna_minimax}
L.~Fox and I.~B. Barker, {\em Chebyshev Polynomials in Numerical Analysis}.
\newblock Oxford University Press, 1968.

\bibitem{rawlings2017model}
J.~Rawlings, D.~Mayne, and M.~Diehl, {\em Model Predictive Control: Theory, Computation, and Design}.
\newblock Nob Hill Publishing, 2017.

\bibitem{borrelli2017predictive}
F.~Borrelli, A.~Bemporad, and M.~Morari, {\em Predictive Control for Linear and Hybrid Systems}.
\newblock Cambridge University Press, 2017.

\bibitem{Notaris}
S.~E. Notaris, ``Interpolary quadrature formulae with chebyshev abscissae,'' {\em Journal of Computational and Applied Mathematics}, vol.~133, pp.~507--517, 2001.

\bibitem{osqp-codegen}
G.~Banjac, B.~Stellato, N.~Moehle, P.~Goulart, A.~Bemporad, and S.~Boyd, ``Embedded code generation using the {OSQP} solver,'' in {\em IEEE Conference on Decision and Control (CDC)}, 2017.

\bibitem{TPODS_system}
D.~Parikh~et al., ``A scalable tabletop satellite automation testbed: Design and experiments,'' in {\em 2023 AAS - Rocky Mountain GN\&C Conference}, 02 2023.

\bibitem{TPODS_GNC24}
D.~Parikh and M.~Majji, ``Pose estimation of cubesats via sensor fusion and error-state extended kalman filter,'' in {\em 2024 AAS - Rocky Mountain GN\&C Conference}, 02 2024.

\bibitem{TPODS_multi_agent}
D.~Parikh, D.~van Wijk, and M.~Majji, ``Safe multi-agent satellite servicing with control barrier functions,'' in {\em 2025 AAS - Rocky Mountain GN\&C Conference}, 02 2025.

\bibitem{DCOL}
K.~Tracy, T.~A. Howell, and Z.~Manchester, ``Differentiable collision detection for a set of convex primitives,'' in {\em 2023 IEEE International Conference on Robotics and Automation (ICRA)}, pp.~3663--3670, 2023.

\end{thebibliography}

\end{document}